\documentclass[11pt,a4paper]{article} \usepackage[hyperref]{acl2017} \usepackage{times} 
\usepackage{latexsym}

\usepackage{url}

\aclfinalcopy 

\title{Duluth at SemEval-2017 Task 7:\\Puns Upon a Midnight Dreary, 
Lexical Semantics for the Weak and Weary}

\author{Ted Pedersen\\
 Department of Computer Science\\
 University of Minnesota\\
 Duluth, MN 55812 USA\\
 {\tt tpederse@d.umn.edu}}

\date{}

\begin{document}
\maketitle
\begin{abstract}
This paper describes the Duluth systems that participated in
SemEval-2017 Task 7 : Detection and Interpretation of English
Puns. The Duluth systems participated in all three subtasks, 
and relied on methods that included word sense
disambiguation and measures of semantic relatedness. 
\end{abstract}

\section{Introduction}

Puns represent a broad class of humorous word play. This paper
focuses on two types of puns, \textit{homographic} and \textit{heterographic}.

A \textit{homographic} pun is characterized by an oscillation 
between two senses of a single word, each of which leads to 
a different but valid interpretation: 

\begin{quote}
I'd like to tell you a chemistry joke but I'm afraid of your reaction.
\end{quote}

Here the oscillation is between two senses of {\em reaction}. The first 
that
comes to mind is perhaps that of a person revealing their true feelings
about something (how they react), but then the relationship to 
{\em chemistry} emerges and the reader realizes that {\em reaction}
can also mean the chemical sense, where substances change into others. 

Homographic puns can also be created via compounding: 
\begin{quote}
He had a collection of candy that was in mint condition. 
\end{quote}
The pun relies on the oscillation between the flavor {\em mint}
and the compound {\em mint condition}, where {\em candy} interacts 
with {\em mint} and {\em mint condition} interacts with {\em collection}.

A \textit{heterographic} pun relies on a different kind of oscillation, 
that is between two words that nearly sound alike, rhyme, or are nearly 
spelled the same. 

\begin{quote}
The best angle from which to solve a problem is the try angle. 
\end{quote}

Here the oscillation is between \textit{try angle} and \textit{triangle}, 
where {\em try} suggests that the best way to solve a problem is 
to try harder, and \textit{triangle} is (perhaps) the best kind of angle. 

This example illustrates one of the main challenges of heterographic 
puns, and that is identifying multi word expressions that are used as a 
kind of compound, but without being a standard or typical compound
(like the very non-standard {\em try angle}). 
One reading treats {\em try angle} as a kind of
misspelled version of {\em triangle} while the other treats them as two
distinct words ({\em try} and {\em angle}). There is also a kind of 
oscillation between senses here, since {\em try angle} can waver back and
forth between the geometric sense and the one of making effort. 

During our informal study of both heterographic and homographic puns, 
we observed a fairly clear pattern
where a punned word will occur towards the end of a sentence and
has a sense that is semantically related to an earlier word,
and another sense that fits the immediate 
context in which it occurs. It often seemed that the sense that
fits the immediate context is a more conventional usage (as in
{\em afraid of your reaction}) and the more amusing sense is
that which connects to an earlier word via some type of semantic
relation ({\em chemical reaction}). This
is more complicated in the case of heterographic puns since
the punned word can rely on pronunciation or spelling
to create the effect (i.e., {\em try angle} versus
{\em triangle}). In this work we focused
on exploiting these long distance semantic relations, although
in future work we plan to consider the use of
language models to identify more conventional usages. 

We used two versions of the WordNet SenseRelate word sense
disambiguation 
algorithm\footnote{\url{http://senserelate.sourceforge.net}} :
TargetWord \cite{PatwardhanBP05} 
 and AllWords \cite{PedersenK09}. Both have
the goal of finding the assignment of senses in a context 
that maximizes their overall semantic relatedness
\cite{PatwardhanBP03} according to measures in
WordNet::Similarity\footnote{\url{http://wn-similarity.sourceforge.net}}
\cite{PedersenPM04a}. We relied on the
Extended Gloss Overlaps measure (lesk) 
\cite{BanerjeeP03b} and the Gloss vector measure 
(vector) \cite{PatwardhanP06}. 

The intuition behind a Lesk measure is that related words will
be defined using some of the same words, and that recognizing
these overlaps can serve as a means of identifying relationships
between words \cite{Lesk86}. The Extended Gloss overlap measure
(hereafter simply lesk) extends this idea by considering not only 
the definitions of the words themselves, but also concatenates
the definitions of words that are directly related via hypernym,
hyponym, and other relations according to WordNet. 

The Gloss Vector measure (hereafter simply vector) extends
this idea by representing each word in a concatenated definition
with a vector of co-occurring words, and then 
creating a representation of this definition by averaging
together all of these vectors. The relatedness between
two word senses can then be measured by finding the
cosine between their respective vectors. 


\section{Systems}

The evaluation data for each subtask was individual sentences
that are 
independent of each other. All sentences were tokenized so 
that each alphanumeric string was separated from any adjacent punctuation, 
and all text was converted to lowercase. Multi-word expressions 
(compounds) found in WordNet were identified. 

SemEval--2017 Task 7 \cite{MillerHG17} focused on pun identification, and was
divided into three subtasks. 

\subsection{Subtask 1}

The problem in Subtask 1 was to identify if a sentence contains a
pun (or not). We relied on the premise that a sentence will
have one unambiguous assignment of senses, and that this should be
true even as the parameters of a word sense disambiguation algorithm
are varied. Thus, if a sentence has multiple possible assignments
of senses based on the results of different runs of a word sense
disambiguation algorithm, then there is a possibility that a pun
exists. To investigate this hypothesis we ran the 
WordNet::SenseRelate::AllWords algorithm using four
different configurations, and then compared the four sense tagged
sentences with each other. If there were more than two differences
in the sense assignments that resulted from these different runs,
then the sentence is presumed to contain a pun.  

WordNet::SenseRelate::AllWords takes measures of semantic relatedness 
between all the pairwise combinations of words in a sentence that 
occur within a certain number of positions of each other (the
window size), and assigns the sense to each 
content word that results in the maximum relatedness among
the words in that window. The assumption that underlies this method is
that words in a window will be semantically related, at least to 
an extent, so when choices among word senses are made, those
that are most related to other words in the window will be
selected. 

The four configurations include two where the window of context
is the entire sentence (a wide window) and another two where the
window of context is only one word to the left and one word
to the right (a narrow window). In addition these two configurations
were carried out with and without compounding of words being
performed prior to disambiguation.
In all four configurations the Gloss Vector measure 
WordNet::Similarity::vector was used as the measure of semantic
relatedness. If more than two sense changes
result from these different configurations, then we say that
a pun has occurred in the sentence. 

\subsection{Subtask 2}

In Subtask 2 the evaluation data consists of the instances from 
Subtask 1 that contain puns. The task is to identify the punning
word. 

We took two approaches to this subtask, however both were informed
by our observation that punned words often occur later
in sentences. The first (run 1) was 
to rely on our word sense disambiguation results from Subtask 1 and
identify the last word which changed senses between different runs
of the WordNet::SenseRelate::AllWords disambiguation algorithm. 
We relied on two of the four 
configurations used in Subtask 1. We used the narrow and wide
contexts from Subtask 1 without finding compounds. We realized 
that this might cause us to miss some cases where a pun was
created with a compound, but our intuition was that the more common 
cases (especially for homographic puns) would be those without 
compounds. Our second approach (run 2) was a simple baseline where the
last content word in the sentence was simply assumed to be the
punned word. 

\subsection{Subtask 3}

The evaluation data for Subtask 3 includes heterographic and homographic
instances from Subtask 2
where the word being punned has been identified. The task is to 
determine which two senses of the punned word are creating the pun.

We used the word sense disambiguation algorithm
WordNet::SenseRelate::TargetWord, which assigns a sense to a single
word in context (whereas AllWords assigns a sense to every word
in a context). 
However, both TargetWord and AllWords have the same
underlying premise, and that is that words in a sentence should
be assigned the senses that are most related to the senses of
other words in that sentence. 

We tried various combinations of TargetWord configurations,
where each would produce their own verdict on the sense of the punned
word. We took the two most frequent senses assigned by these variations
and used them as the sense of the punned word. Note that for the
heterographic puns there was an additional step, where alternative
spellings of the target word were included in the disambiguation algorithm.
For example :
\begin{quote}
The dentist had a bad day at the orifice. 
\end{quote}
{\em Orifice} is already identified as the punned word, and one of 
the intended senses would be that of an opening, but the other is the 
somewhat less obvious spelling variation {\em office}, as in {\em
a bad day at the office}. 


For the first variation (run 1) we used both the local and global options from 
TargetWord. The local option measures the semantic relatedness of the target
word with all of the other members of the window of context, whereas the global option
measures the relatedness among all of the words in the window of context (not just
the target word). We also varied whether the lesk or vector measure was
used, if a narrow or wide window was used, and if compounds were identified. 
We took all possible combinations
of these variations, which resulted in 16 possible configurations.  To this 
we added a WordNet sense one baseline with and without finding compounds, and a
randomly assigned sense baseline. Thus, there were 19 variations in our run 1
ensemble. We took this approach with both the homographic and heterographic 
puns, although for the heterographic puns we also replaced the target word 
with all of the words known to WordNet that differed by one edit distance. 
The premise of this was to detect minor misspellings that might
enable a heterographic pun. 

For run 2 we only used the local window of context with WordNet::SenseRelate::TargetWord, 
but added to lesk and vector the Resnik measure (res) and the shortest path (path) measure.
We carried out each of these with and without identifying compounds,
which gives us a total of eight different combinations. We also tried a much more
ambitious substitution method for the heterographic puns, where we queried the
Datamuse API in order to find words that were rhymes, near rhymes, homonyms,
spelled like, sound like, related, and means like words for the target word. This
created a large set of candidate target words, and all of these were disambiguated
to find out which sense of which target word was most related to the surrounding
context. 


\section{Results}

We review our results in the three subtasks in this section. 
Table \ref{table:subtasks} refers to homographic results as {\em hom}
and heterographic as {\em het}. Thus the first run of the Duluth systems on
homographic data is denoted as Duluth-hom1, and the first run on
heterographic data is Duluth-het1. The highest ranking system is 
indicated via High-hom and High-het. P and R as column headers stand
for precision and recall, A stands for accuracy, and C is for coverage.
Rank x/y indicates that this system was ranked x of y participating systems.

\subsection{Subtask 1}

Puns were found in 71\% (1,271) of the heterographic and 71\% of the 
homographic instances (1,607). This suggests this subtask would have
a relatively high baseline performance, for example if a system 
simply predicted that every sentence contained a pun. Given this
we do not want to make too strong a claim about our approach, but it
does seem that focusing on sentences that have multiple possible (and
valid) sense assignments is promising for pun identification. Our
method tended to over-predict puns, reporting that a pun occurred in
84\% (1,489 of 1,780 instances) of the heterographic data, and 80\% 
(1,791 of 2,250 instances) of the homographic. 

\begin{table}
\begin{center}
\caption{Subtask 1, 2, 3 results}
\label{table:subtasks}
\begin{tabular}{lccccc}
\hline
{\bf Subtask 1}    	& P  	& R  	& A  	& F1  	& rank \\
High-hom	& .97   & .80   & .84   & .87  & 1 / 9 \\
Duluth-hom1   	& .87   & .78   & .74   & .83  & 2 / 9\\
High-het    	& .87   & .82   & .78   & .84  & 1 / 7 \\
Duluth-het1	& .87   & .74   & .69   & .80  & 3 / 7\\ \hline
{\bf Subtask 2}	& P  	& R  	& C  	& F1  	& rank \\
High-hom       & .66   & .66   & 1.0 & .66  & 1 / 15 \\
Duluth-hom1   & .37   & .36   & .99   & .37  & 7 / 15 \\
Duluth-hom2   & .44   & .44   & 1.0   & .44  & 6 / 15 \\
High-het       & .80   & .80   & 1.0   & .80  & 1 / 11 \\
Duluth-het1   & .18   & .18   & .99   & .18  & 11 / 11 \\
Duluth-het2   & .53   & .53   & 1.0   & .53  & 4 / 11 \\ \hline
{\bf Subtask 3} 	& P  	& R  	& C  	& F1  	& rank \\
High-hom       & .17   & .14   & .86   & .16  & 1 / 8 \\
Duluth-hom2   & .17   & .14   & .86   & .16  & 1 / 8 \\
Duluth-hom1   & .15   & .15   & 1.0   & .15  & 3 / 8 \\
High-het     & .08   & .07   & .83   & .08  & 1 / 6 \\
Duluth-het1 & .03   & .03   & 1.0   & .03  & 3 / 6 \\
Duluth-het2 & .001   & .001   & .98   & .001  & 6 / 6 \\
\hline
\end{tabular}
\end{center}
\end{table}

\subsection{Subtask 2}

Subtask 2 consists of all the instances from Subtask 1 that 
included a pun. This leads to 1,489 heterographic puns and 1,791 homographic. 

We see that our simple baseline method of choosing the last content
word as the punned word (run 2) significantly outperformed our  more
elaborate method (run 1) of identifying which word experienced more changes
of senses across multiple variations of the disambiguation algorithm.
We can also see that run 1 did not fare very well with heterographic
puns. In general we believe the difficulty that run 1 experienced was
due to the overall noisiness that is characteristic of word sense
disambiguation algorithms.

\subsection{Subtask 3}

Subtask 3 consists of 1,298 homograph instances and 1,098 heterographic 
instances. We see that for homographs our method fared very well, and 
was the top ranked of participating systems. On the other hand our 
heterographic approach was not terribly successful. We believe
that the idea of generating alternative target words for heterographic
puns is necessary, since without this it would be impossible to identify
one of the senses of the punned word. 
However, our run 1 approach of simply using target word variations
with an edit distance of one did not capture the variations present in 
heterographic puns (e.g., {\em orifice} and {\em office} have an
edit distance of 2). Our run 2 approach of finding many different
target words via the Datamuse API resulted in an overwhelming
number of possibilities where the intended target word was very 
difficult to identify. 

\section{Discussion and Future Work}

One limitation of our approach is the uncertain level 
of accuracy of word sense disambiguation algorithms, which vary
from word to word and domain to domain.
Finding multiple possible senses for a single word
may signal a pun or expose the limits of
a particular WSD algorithm. 

In addition, the contexts used in this evaluation were all
single sentences, and were relatively short. Whether or 
not having more context available would help or hinder these approaches
is an interesting question.

Heterographic puns posed a host of challenges, in particular mapping
clever near spellings and near pronunciations into the intended form
(e.g., {\em try angle} as {\em triangle}). Simply trying to assign 
senses to {\em try angle} will obviously miss the pun, and so the
ability to map similar sounding phrases to the intended word is a 
capability that our systems were not terribly successful with. However,
we were better able to identify compounds in homographic puns
(e.g., {\em mint condition}) since those were written literally
and could be found (if in WordNet) via a simple subsequence
search. 

While our reliance on word sense disambiguation
and semantic relatedness served us well for homographic puns,
it was clearly not sufficient for heterographic. Moving forward
it seems important to have a reliable mechanism to map the
spelling and pronunciation variations that characterize 
heterographic puns to their intended forms. While dictionaries
of rhyming and sound-alike words are certainly helpful, they
typically introduce too many possibilities from which to make
a reliable selection. Language modeling seems like a promising
way to winnow that space, so that we can
get from a {\em try angle} to a {\em triangle}. 
      

\end{document}